\documentclass[letterpaper, 10 pt, conference]{ieeeconf}  % Comment this line out if you need a4paper

\IEEEoverridecommandlockouts                              % This command is only needed if 
\usepackage{graphics} % for pdf, bitmapped graphics files
\usepackage{epsfig} % for postscript graphics files
\usepackage{mathptmx} % assumes new font selection scheme installed
\usepackage{times} % assumes new font selection scheme installed
\usepackage{amsmath} % assumes amsmath package installed
\usepackage{amssymb}  % assumes amsmath package installed

\title{\LARGE \bf
A Survey Forest Diagram : Gain a Divergent Insight View on a Specific Research Topic *
}

\author{JingHong Li$^{1}$, Wen Gu$^{2}$, Koichi Ota$^{2}$ and Shinobu Hasegawa$^{2}$% <-this % stops a space
\thanks{*This work was supported by JSPS KAKENHI Grant
Number JP20H04295.}% <-this % stops a space
\thanks{$^{1}$Division of Advanced Science and Technology, Japan Advanced Institute of Science and Technology, Asahidai, Nomi, 9231292, Ishikawa, Japan.}%
\thanks{$^{2}$Center for Innovative Distance Education and Research, Japan Advanced Institute of Science and Technology, Asahidai, Nomi, 9231292, Ishikawa, Japan.}%
}

\begin{document}

\maketitle
\thispagestyle{empty}
\pagestyle{empty}

\begin{figure*}[htbp]
\centering
\fbox
{
\includegraphics[width=17.0cm,height=7.0cm]{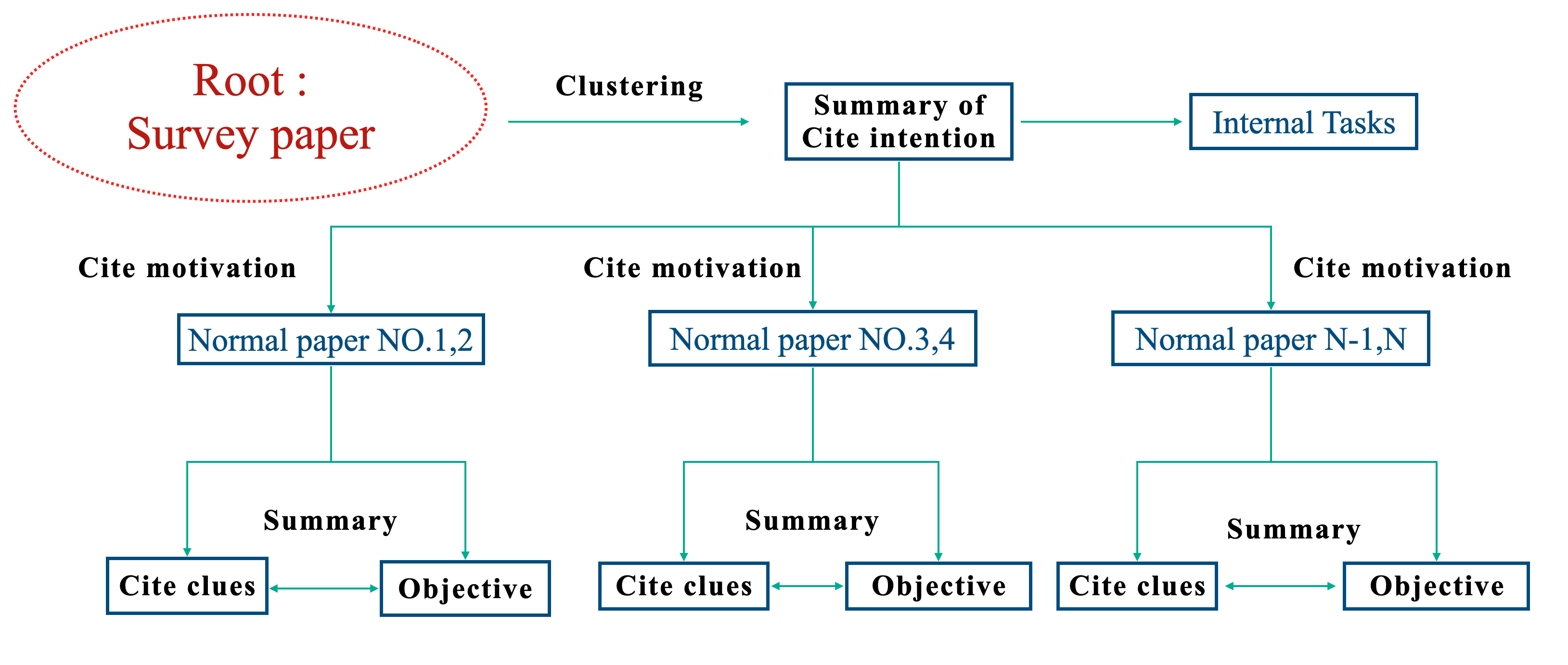}
}
\caption{Overview of the strategy}
\label{Figure_Strategy}
\end{figure*}

%%%%%%%%%%%%%%%%%%%%%%%%%%%%%%%%%%%%%%%%%%%%%%%%%%%%%%%%%%%%%%%%%%%%%%%%%%%%%%%%
\begin{abstract}
With the exponential growth in the number of papers and the trend of AI research, the use of Generative AI for information retrieval and question-answering has become popular for conducting research surveys. However, novice researchers unfamiliar with a particular field may not significantly improve their efficiency in interacting with Generative AI because they have not developed divergent thinking in that field. This study aims to develop an in-depth Survey Forest Diagram that guides novice researchers in divergent thinking about the research topic by indicating the citation clues among multiple papers, to help expand the survey perspective for novice researchers. 
\end{abstract}

%%%%%%%%%%%%%%%%%%%%%%%%%%%%%%%%%%%%%%%%%%%%%%%%%%%%%%%%%%%%%%%%%%%%%%%%%%%%%%%%

\vspace{-1.0mm}
\section{Introduction}
\label{main}
\vspace{-1.0mm}
With the widespread use of Generative AI, novice researchers tend to use \textit{ChatGPT}-based interfaces for academic Q\&A to efficiently understand the background knowledge and key points of their research topics\cite{fish-bone}. Recently, AI tools for education have used advanced algorithms to evaluate large amounts of data, providing personalized and adaptable research survey experiences, such as \textit{chatpdf}, \textit{scispace}, \textit{paper-connection}, \textit{ResearchRabbit}, \textit{Consensus}, \textit{Elicit} \cite{AI-Tools}. However, due to the novice researcher's lack of experience and advanced training, they often struggle to ask effective questions based on keywords to get the desired answers. The main reason is that they do not understand the origin of the research issue, making it difficult to link foundational knowledge of the research task. To solve this issue, previous research has tried to extract implicit associative information from multiple papers by analyzing why authors use citations\cite{funding-and-citation}\cite{tree-kg}. However, these studies did not systematically analyze the intentions, motivations, and clues behind the citations. Hence, novice researchers struggle to construct logical structures from extensive citation information, affecting the efficiency of their surveys. This study aims to assist novice researchers understand the divergent insights from multiple papers on the same research topic. Starting with a survey paper as the survey root, it creates a divergent-thinking forest diagram based on three characteristics of citations: intentions, motivations, and clues, to help expand the survey perspective for novice researchers.

\section{Methodology}
 We try to discover divergent directions of survey papers by defining the characteristics of their citations. We create multiple layers and embed the corresponding summaries that reflect these citation characteristics in each layer. Finally, we construct a survey forest diagram by linking survey papers, cited normal papers, and their summaries of the citation characteristics. The definition of Divergent Insight Survey and its internal concepts shows as follows:\\
\textbf{(1) Divergent Insight Survey: }Divergent thinking is a cognitive style that facilitates idea generation in situations with vague selection criteria and multiple correct solutions, emphasizing mental flexibility\cite{divergent}. In academic surveys, insights from divergent guidance can offer novice researchers helpful hints. This enables them to conduct more in-depth surveys from various perspectives and directions.\\
\textbf{(2) Survey Forest Diagram: }The forest diagram displayed results from paired observations and events for the similar article of feature, along with overall effects\cite{forest}. In a research topic, there are multiple survey papers, and each survey paper cites multiple normal papers. They may have related research purposes and directions, and the characteristics of the forest diagram will adapt to the expression of such multi-directional, multi-branch overall effects.\\
\textbf{(3) Survey Paper \& Normal Paper: }A survey paper, is organized by experts in the field to provide background knowledge, related tasks, and future direction speculations. It systematically presents an overview of specific research topic. A survey paper can be considered a root. By combining this root with the papers cited in various section, we can connect their citation logic to form a citation clue.\\
\textbf{(4) Citation intention: }The authors of a survey paper conduct extensive literature research on a particular topic and organize the documents into sections based on research task segmentation. Hence, the citations in each section embed the author's intention, reflecting the author's direction of citation.\\
\textbf{(5) Citation motivation: } The authors of survey papers usually indicate their motivation in the in-text citations. Similar to citation sentiment, this explains why the author cites a particular paper\cite{important-citation}. The text embedded in and near the citation often reflects the author's citation motivation. \\
\textbf{(6) Citation clues: } To determine the connection between a cited work and the citing document, analyze clues from the latter’s author. Deep-mining on the citation’s purpose, function, and motive is essential for measuring the work’s impact\cite{cite-clues}. In this study, a citation clue consists of citation intention and citation motivation. Creating a summary that reflects the citation clue can show the logical structure between the survey paper and the papers it cites more intuitively. This assists novice researchers to logically diverge on key research points.

Based on the above concepts, we try to integrate a strategy to support the Divergent Insight Survey in multiple layers. We try to demonstrate the Divergent Insight View through the following process: \textbf{(1)} First, we extract survey papers from the 'HotpotQa' topic in the \textit{S2orc} dataset, along with the normal papers they cite, to form the basic prototype (nodes and edges) of the forest diagram. \textbf{(2)} Next, we locate the citation content based on the section where a normal paper is cited in a survey paper and pinpoint the location of the citation (which sentence is cited in the text). To make the diagram structure more concise, we cluster the citation content to present multiple tasks based on the citation content and expand into the citation intent. \textbf{(3)} Then, we use prompt engineering combined with citation content to create abstractive summarization that highlights the citation clues of normal papers. This showcases the citation clues of each normal paper within the entire forest, forming the extension nodes of each paper. \textbf{(4)} Finally, we extend another edge from the normal paper within each diagram to output its research objective, and provides a summary of the association between the cite clue and the research objective, specifically expressing the connection between survey papers and normal papers. The entire process is implemented using prompt engineering.
The strategy overview is shown in Figure \ref{Figure_Strategy}.

\section{Conclusion}
This study specifically defines the direction of the Divergent Insight Survey and conducts initial framework development. The goal of this study is to develop an in-depth Survey Forest Diagram that guides novice researchers in divergent thinking about the research topic by indicating the citation intentions among multiple papers, enabling them to quickly gain insights into potential research elements. Further development will consider the following directions:\\
\textbf{(1) Constructing a more accurate prompt description: }To generate citation clues that match the characteristics of novice researchers, we need to design and develop more refined prompt templates to ensure that the generated content of meets our expectations.\\
\textbf{(2) Adjustable insight range template: }Due to the excessive number of papers on the topic, providing numerous summaries may still confuse novice researchers. Therefore, we will design an Adjustable Insight Range Template. This template functions to optimize the scale and amount of information in the diagram based on user input parameters.\\
\textbf{(3) Adjuster for summary generation: }In addition to controlling the scope of the diagram, the length of the summary should balance readability and sufficient to improve the survey efficiency of novice researchers.\\

\end{document}